\documentclass[runningheads]{llncs}

\usepackage[T1]{fontenc}
\usepackage{graphicx}
\usepackage{hyperref}
\usepackage{color}

\usepackage{booktabs}
\usepackage{multirow}

\begin{document}
\title{Self-Ensembling Vision Transformer (SEViT) for Robust Medical Image Classification}
%
\titlerunning{SEViT}
%
\author{Faris Almalik  
\and
Mohammad Yaqub 
 \and
 Karthik Nandakumar
}
\authorrunning{F. Almalik et al.}
 \institute{Mohamed Bin Zayed University of Artificial Intelligence, Abu Dhabi, UAE \email{\{faris.almalik, mohammad.yaqub, karthik.nandakumar\}@mbzuai.ac.ae}}
 
\maketitle              
\begin{abstract}
Vision Transformers (ViT) are competing to replace Convolutional Neural Networks (CNN) for various computer vision tasks in medical imaging such as classification and segmentation. While the vulnerability of CNNs to adversarial attacks is a well-known problem, recent works have shown that ViTs are also susceptible to such attacks and suffer significant performance degradation under attack. The vulnerability of ViTs to carefully engineered adversarial samples raises serious concerns about their safety in clinical settings. In this paper, we propose a novel self-ensembling method to enhance the robustness of ViT in the presence of adversarial attacks. The proposed Self-Ensembling Vision Transformer (SEViT) leverages the fact that feature representations learned by initial blocks of a ViT are relatively unaffected by adversarial perturbations. Learning multiple classifiers based on these intermediate feature representations and combining these predictions with that of the final ViT classifier can provide robustness against adversarial attacks. Measuring the consistency between the various predictions can also help detect adversarial samples. Experiments on two modalities (chest X-ray and fundoscopy) demonstrate the efficacy of SEViT architecture to defend against various adversarial attacks in the gray-box (attacker has full knowledge of the target model, but not the defense mechanism) setting. Code: \url{https://github.com/faresmalik/SEViT} 

\end{abstract}
\section{Introduction}

Convolutional Neural Networks (CNNs) have been the de facto models for medical image analysis tasks such as segmentation \cite{Segmentation}, landmark localization \cite{localization}, and classification \cite{classification}. However, it is well-known that CNNs trained on natural or medical images are vulnerable to adversarial attacks \cite{attack_CNN_2,FGSM}, which add imperceptible perturbations to input images to deliberately mislead a target model. The work in \cite{incentives} has identified two weak links in the healthcare economy that are susceptible to adversarial attacks. Firstly, automated systems deployed by insurance companies to process reimbursement claims can be fooled using adversarial samples to trigger specific diagnostic codes and obtain higher payouts. Secondly, automated systems deployed by regulators to confirm results of clinical trials can be circumvented by malicious manufacturers, who can employ adversarial test samples to pass clinical trials successfully without being noticed. In addition, the rapid growth of telemedicine during the COVID-19 pandemic and the emergence of “as-a-service” business models based on cloud computing (e.g., radiology-as-a-service) have created an environment where medical images will be increasingly processed remotely. Often, automated machine learning algorithms will perform the diagnosis, which human experts may optionally verify. However, such medical imaging scenarios will be highly vulnerable to adversarial attacks. Thus, a robust defensive strategy must be devised before automated medical imaging systems can be securely deployed.

\begin{figure}[t]
\centering
\includegraphics[width=0.85\textwidth]{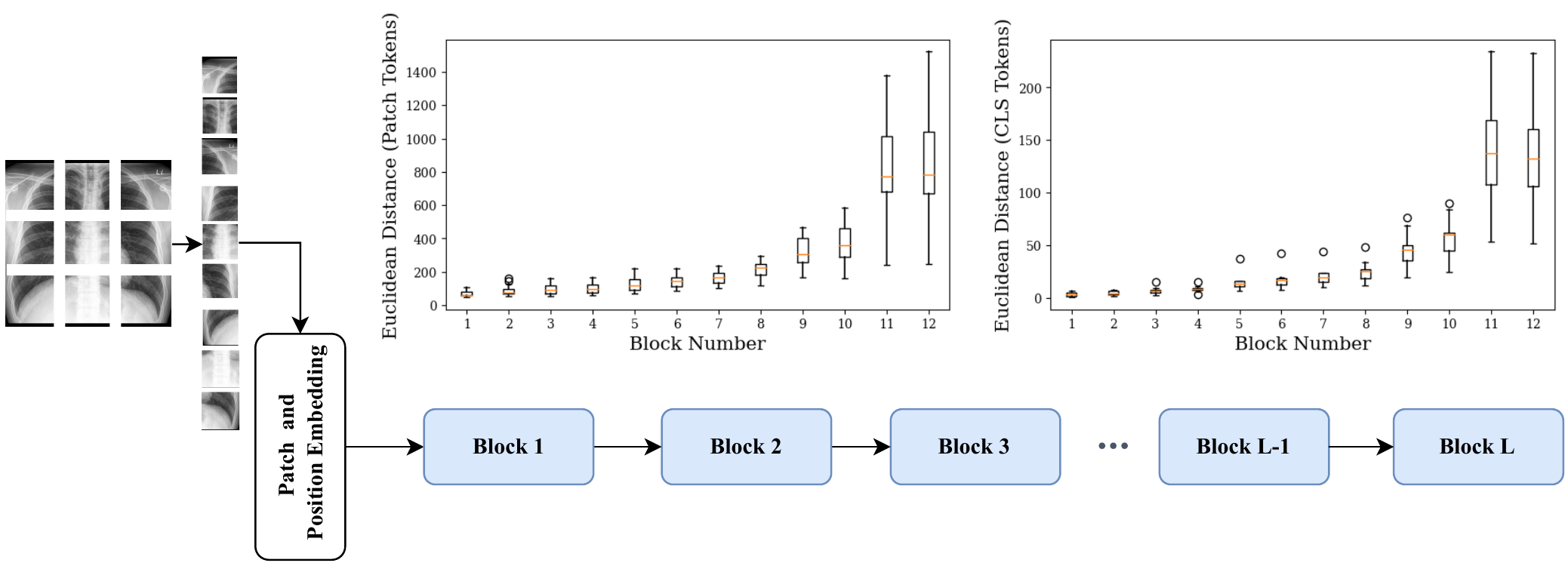}
\caption{Euclidean distance between class and patch tokens of adversarial images and the corresponding clean version. The distance increases moving towards the final blocks.}
\label{featuredistance}
\end{figure}

Recently, self-attention-based neural network architectures such as Vision Transformers (ViT) \cite{viT} have proven to be successful in image classification. Consequently, ViTs are competing to replace CNNs in various tasks related to medical images \cite{detection,classification-medic-trans,segmentation-med-transform}. Despite the superior performance of ViTs, recent works have confirmed that they are also susceptible to malicious perturbations \cite{vulner_transf1,vulner_transf3}. ViTs typically partition the input image into multiple patches and learn feature representations (called \emph{patch tokens}) for each patch as they pass through multiple Transformer blocks. In addition, each block distills information from the patch tokens into a global representation (called \emph{class token}). The class token output by the final Transformer block is often used for classification. 

Adversarial attacks can be expected to push the feature representations of the perturbed samples away from that of the clean samples. However, a careful analysis of ViTs indicates that the impact of perturbations is more pronounced in the final set of blocks and the patch tokens learned by the initial blocks remain relatively unaffected (see Fig. \ref{featuredistance}). This phenomenon raises two main questions: (i) are the features learned by initial blocks useful for classification? and (ii) can we use these intermediate features to enhance the robustness of ViTs? 

We propose a novel method that utilizes the patch tokens learned by initial blocks in ViTs along with the final class token to enhance the classification robustness and detect adversarial samples. To the best of our knowledge, this is the first work that enhances the robustness of ViTs against adversarial attacks in medical imaging classification. The contributions of this paper are: (i) we propose a self-ensembling method to enhance the adversarial robustness of ViT for medical image classification and (ii) we propose an adversarial sample detection method based on the consistency of predictions made by the classifiers in the ensemble.

\subsubsection{Related Work.}
The use of ViTs in medical imaging has seen a rapid increase recently \cite{TransformersInMedicalImagingSurveyPaper}. Several studies have shown that off-the-shelf ViTs are vulnerable to adversarial perturbations \cite{FGSM,PGD,BIM}. However, they were found to be more robust against adversarial attacks compared to CNNs \cite{shao,vulner_transf1,Nasseretal_NeurIPS2021}. Low transferability of adversarial samples between CNNs and ViTs has also been observed \cite{vulner_transf3,Nasseretal_ICLR2022}. While these initial results were promising, recent works have demonstrated that it is indeed possible to fool ViTs using adversarial samples. For example, the Patch-Fool attack proposed in \cite{patch-fool} targets the self-attention mechanisms in ViTs making them weaker learners compared to CNNs. Similarly, an ensemble (of CNN and ViT) defense approach was found to be ineffective against white-box adversaries \cite{vulner_transf3}. Furthermore, Nasser et al. \cite{Nasseretal_ICLR2022} show that it is possible to generate more powerful adversarial attacks against ViTs by targeting an ensemble of intermediate representations in addition to the final class token.

Many defense mechanisms have been proposed to improve the robustness of machine learning models against adversarial attacks. These include adversarial training \cite{FGSM,PGD,adv_training1}, defensive distillation \cite{Papernotetal_SP2016} and adversarial purification \cite{DefenseGAN}. Algorithms have also been proposed to detect and filter out adversarial samples \cite{KDE,LID,MagNet}. However, almost all these defense mechanisms have been designed for CNNs and there is limited work on enhancing the adversarial robustness of ViTs. The most notable approaches are PatchVeto \cite{PatchVeto} and Robust Self-Attention \cite{RSA_MuWagner_UDL_ICML2021}, both of which are designed to defend ViTs against adversarial patch attacks.

\section{Proposed Method}
\begin{figure}[t!]
\centering
\includegraphics[width=0.83\textwidth]{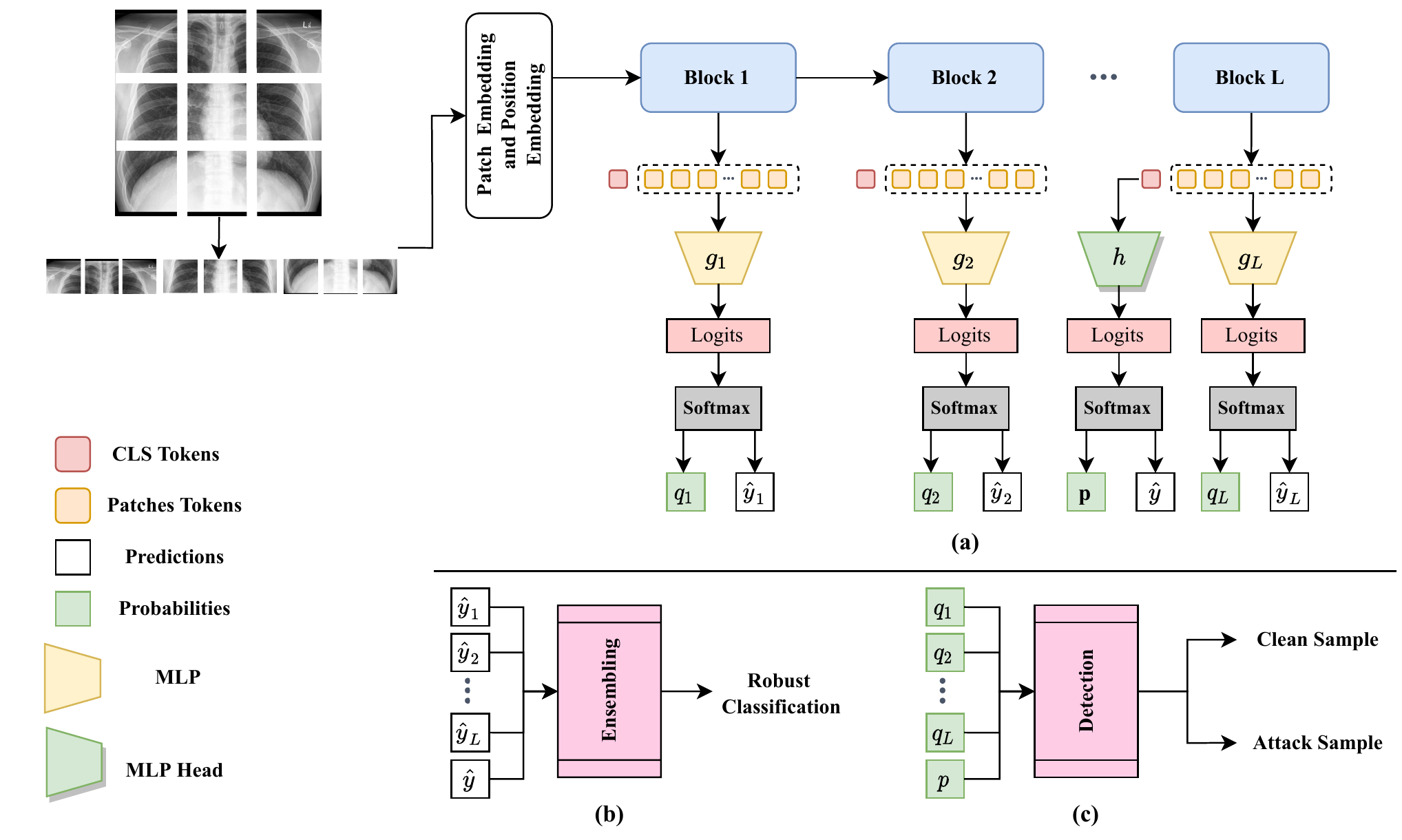}
\caption{The proposed SEViT framework extracts the patch tokens from the initial blocks and trains separate MLPs as shown in (a). (b) A self-ensemble of these MLPs with the final ViT classifier enhances the robustness of ViT. (c) Consistency between the predictions in the ensemble can be used to detect adversarial samples.}
\label{method}
\end{figure}

Let $\mathbf{x}$ be a medical image with true label $y$ provided as input to a ViT-based classifier $\mathbf{f}(\cdot)$. The goal of an adversarial attack is to craft a perturbed image $\mathbf{x}'$ such that $\mathbf{f}(\mathbf{x}')\neq y$ with high probability under the constraint that $\mathbf{x}'$ is \emph{close} to $\mathbf{x}$. In this work, we use $L_\infty$ norm as the distance metric and ensure that $||\mathbf{x}-\mathbf{x}'||_\infty \leq \epsilon$. The objective of our defense mechanism is to obtain a robust classifier $\tilde{\mathbf{f}}$ from $\mathbf{f}$ such that both the clean accuracy ($P(\tilde{\mathbf{f}}(\mathbf{x}) = y)$) and robust accuracy ($P(\tilde{\mathbf{f}}(\mathbf{x}') = y)$) are high. The aim of the detection mechanism is to discriminate between $\mathbf{x}$ and $\mathbf{x}'$ with high accuracy, especially when the attack is successful ($\mathbf{f}(\mathbf{x}')\neq y$).

Vanilla ViTs divide the input image $\mathbf{x}$ into $N$ patches of size $P \times P$ each. Patches are then flattened into an one-dimensional (1D) vector and embedded into $D$ dimensions via a linear layer. Since patch embeddings do not preserve the positional information, a 1D learnable position embedding is added to create $N$ patch tokens ($\{pt_0^j\}_{j=1}^N$). Furthermore, a separate class token ($ct_0$) is added to consolidate information from all the patch tokens. These patch and class tokens get refined as they pass through $L$ Transformer blocks. Let $\{pt_i^j\}_{j=1}^N$ and $ct_j$ be the patch and class tokens (respectively) output by the $i^{th}$ block, $i=1,2,\cdots,L$. The final class token $ct_L$ output by the $L^{th}$ block is passed through a multi-layer perceptron (MLP) classification head ($\mathbf{h}_{\theta}$) to obtain the final classification result, i.e., $\mathbf{f}(\mathbf{x})=\mathbf{h}_{\theta}(ct_L)$. The MLP typically includes a final softmax layer to produce a probability distribution $\mathbf{p}$ over the possible classes.

\subsection{Self-Ensembling for Robust Classification}
Vanilla ViTs rely only on the final class token $ct_L$ for image classification and ignores the rest of the learned features. As illustrated in Fig. \ref{featuredistance}, adversarial attacks have a significant impact on the patch and class tokens learned by the later blocks, while the corresponding tokens produced by the initial blocks remain relatively robust. Our main hypothesis is that the intermediate feature representations output by the initial blocks are useful for classification and harder to attack. Hence, we propose to add a MLP classifier at the end of each block. Each intermediate MLP classifier (denoted as $\mathbf{g}_{\beta_i}$, $i=1,2,\cdots,(L-1)$) utilizes the patch tokens of the corresponding block to produce a probability distribution $\mathbf{q}_i$ over the class labels. This results in a self-ensemble of $L$ classifiers that can be fused to obtain the final classification result. We refer to this architecture as Self-Ensembling Vision Transformer (SEViT), which is illustrated in Fig. \ref{method}.

Unlike the final class token $ct_L$, we observed that the intermediate class tokens are not discriminative enough. In contrast, the intermediate patch tokens contained useful information. This is the reason for constructing the intermediate MLPs $\mathbf{g}_{\beta_i}$ based on patch tokens rather than the class token. Moreover, we observe that it is sufficient to add MLP classifiers only to the first $m$ ($m < L$) blocks (see Fig. \ref{MLPs_Clean_accuracy_both_data} (b) and (c)), which are more robust to adversarial attacks, and combine their results with the final classification head. This reduces computational complexity and increases adversarial robustness. Thus, the SEViT model can be expressed as:
\begin{equation}
    \label{SEvIT}
    \tilde{\mathbf{f}}(\mathbf{x}) = \mathcal{F}\left(\mathbf{g}_{\beta_1}(\{pt_1^j\}_{j=1}^{N}),\cdots,\mathbf{g}_{\beta_m}(\{pt_m^j\}_{j=1}^{N}),\mathbf{h}_{\theta}(ct_L)\right)
\end{equation}
where $\mathcal{F}$ denotes the fusion operator, $\theta$ denotes the parameters of the final ViT classifier $\mathbf{h}$, and $\beta_1$ through $\beta_m$ denote the parameters of the $m$ intermediate classifiers $\mathbf{g}$. The outputs of the $(m+1)$ classifiers in the SEViT ensemble can be fused in a number of ways. Let $\{\hat{y}_1,\cdots,\hat{y}_m,\hat{y}\}$ be the class predictions of the $m$ intermediate MLPs and the final ViT classifier. One possibility is to perform majority voting of these $(m+1)$ predictions to obtain the final prediction.
The SEViT ensemble could also be formed by randomly choosing only $c$ out of the initial $m$ intermediate classifiers, $c \in \{1,\dots, m-1\}$ and decisions can be made based on the $(c+1)$ classifier ensemble (including the final ViT classifier). This random selection approach can be expected to be more robust against white-box adversaries, who have full knowledge of the ViT model and intermediate MLPs. 

\subsection{Adversarial Sample Detection}
As noted earlier, adversarial attacks are expected to adversely affect the predictions of the final ViT classifier, whereas the intermediate classifiers at the initial blocks are relatively unaffected. Hence, the predictions of the intermediate classifiers can be expected to be different from that of the final ViT classifier, when an adversarial sample is presented. In contrast, the predictions of all classifiers in SEViT are expected to be in agreement for a clean sample. This phenomenon can be leveraged to detect an adversarial sample in the following way. Let $\mathbf{A}$ be a $((m+1) \times (m+1))$ matrix containing the Kullback-Leibler divergence ($D_{KL}$) between the probability distributions output by the $(m+1)$ classifiers in SEViT. Clearly, the diagonal elements of this matrix will always be zero. For a clean sample, all the non-diagonal entries in this matrix are also expected to be close to zero. On the other hand, at least some of the non-diagonal entries in this matrix (especially elements involving the final ViT classifier) are expected to be large for the adversarial sample. Hence, we compute Frobenius norm of the KL-matrix as follows: 
\begin{equation}
\label{frob_norm}
	\|\mathbf{A}\|_F = \sqrt{\sum_{i=1}^{m+1} \sum_{j=1}^{m+1} |a_{i,j}|^{2}}\:, 
\end{equation}
\noindent $a_{i,j}=D_{KL}(\mathbf{q}_i,\mathbf{q}_j),~\forall~i,j=1,2,\cdots,m$ and $a_{i,m+1}=D_{KL}(\mathbf{q}_i,\mathbf{p}),~\forall~i=1,2,\cdots,m$, and $a_{m+1,m+1}=0$. The Frobenius norm of the KL-matrix is compared to a threshold $\tau$ and the input is detected as adversarial if $\|\mathbf{A}\|_F > \tau$.

\section{Result and Discussion}
\subsubsection{Datasets.} We conduct our experiments on two medical imaging datasets. The first dataset \cite{Tuberculosis_data} consists of $7,000$ chest X-ray images and the classification task is binary, Normal or Tuberculosis. We randomly split $80\%$ of the dataset for training, $10\%$ for validation, and $10\%$ for testing. The second dataset is APTOS2019 \cite{aptos2019} for diabetic retinopathy (DR). It has $5$ classes and $3,662$ retina images. We again convert it to a binary classification task by labeling the images as DR or Normal and randomly select $80\%$ of the images for training and the rest for testing. We refer to this dataset as Fundoscopy.

\subsubsection{Vision Transformer and MLPs.} 
\begin{figure}[t]
\centering
\includegraphics[width=0.9\textwidth]{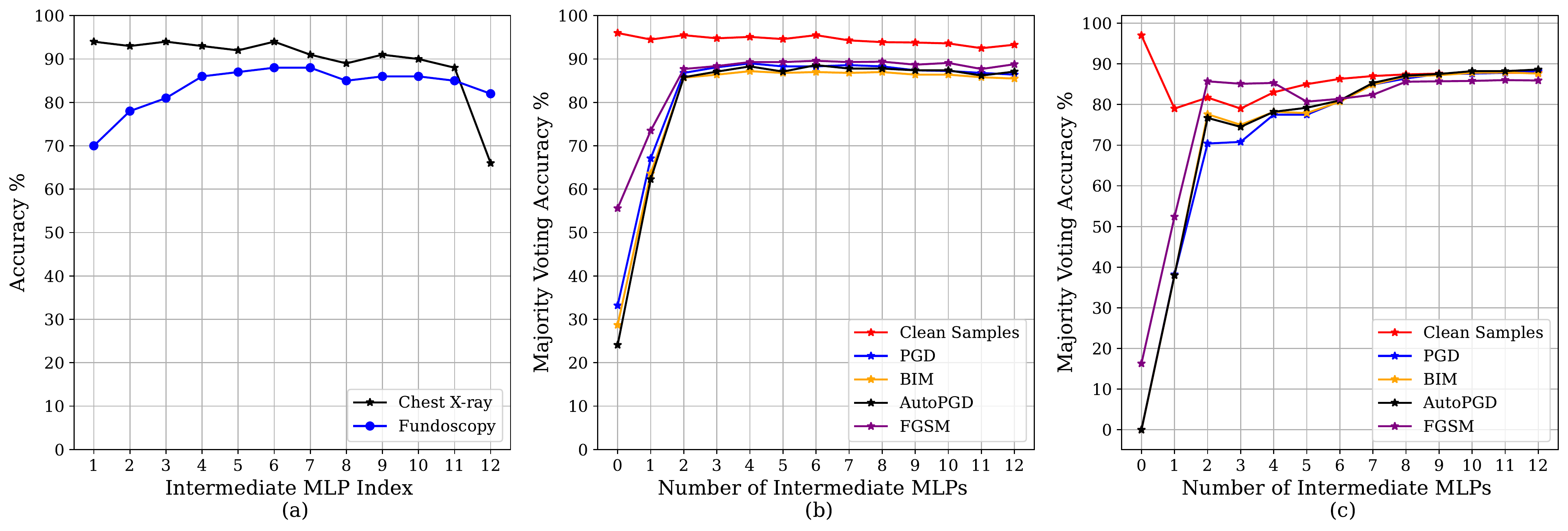}
\caption{ Accuracy ($\%$) of (a) intermediate MLPs on clean test set for both datasets. (b) majority voting from different number of MLPs for chest X-ray dataset. (c) majority voting from different number of MLPs for Fundoscopy dataset}
\label{MLPs_Clean_accuracy_both_data}
\end{figure}

For ViT, we choose the ViT-B/16 model pretrained on ImageNet \cite{ImageNet}. For the intermediate classifiers, we design a 4-layer MLP, which takes the patch tokens as input. We train $12$ separate MLPs, one after each block. Adam optimizer was used with $10^{-3}$ as the initial learning rate and a decay of $0.1$ every $10$ epochs. Different augmentations including random rotation, color jitter, random horizontal and vertical translation are applied during ViT fine-tuning. ViT achieved $96.38\%$ and $97.64\%$ accuracy on the original clean test set for chest X-ray and Fundoscopy, respectively. The accuracy of intermediate classifiers on the original clean test set is depicted in Fig.~\ref{MLPs_Clean_accuracy_both_data} (a). We conduct our experiments using one Nvidia RTX 6000 GPU with 24 GB memory.
\subsubsection{Adversarial Attacks.} We use the Foolbox library \cite{foolbox} to generate various $L_\infty$ attacks such as FGSM \cite{FGSM}, BIM \cite{BIM} and PGD \cite{PGD},  and $L_2$ C\&W \cite{CW} attacks. The attacks are generated on the original test set in a gray-box setting (adversary has full access to the fine-tuned ViT, but is not aware of the intermediate classifiers). For FGSM, PGD, BIM and AutoPGD \cite{AutoPGD} attacks, we generate samples with $\epsilon = 0.003,0.01,0.03$, whereas for C\&W attacks we set the Lagrange multiplier to $2$ and limit the number of steps to 4000. All other parameters are set to default values.

\subsection{Results on Adversarial Robustness of SEViT} 
We compare the clean and robust accuracy of vanilla ViT and SEViT with majority voting based on $m=5$ intermediate classifiers. Quantitative results in Table~\ref{Majoority,allattacks,bothdataset_latest} show that vanilla ViTs are highly susceptible to various adversarial attacks, especially for higher values of perturbation budget $\epsilon$ as can be observed from the drastic fall in robust accuracy (Rows 1 and 3 of Table~\ref{Majoority,allattacks,bothdataset_latest}). On the other hand, SEViT reduced the clean accuracy by $2\%$ and $12\%$ for chest X-ray and Fundoscopy datasets, respectively. However, the robust accuracy of SEViT is significantly higher compared to that of vanilla ViTs for both datasets. In fact, SEViT boosts the adversarial robustness and attains very high robust accuracy across all attacks and for various perturbation budgets (Rows 2 and 4 of Table~\ref{Majoority,allattacks,bothdataset_latest}).
\begin{table}[t]
\centering
\caption{Classification accuracy (\%) of vanilla ViT and SEViT on clean and  adversarial samples with different perturbation budgets.}
\label{Majoority,allattacks,bothdataset_latest}
\resizebox{0.9\textwidth}{!}{%
\begin{tabular}{@{}cccccclccclccclccclc@{}}
\toprule
 &
   &
  Clean Accuracy &
  \multicolumn{3}{c}{FGSM} &
   &
  \multicolumn{3}{c}{PGD} &
   &
  \multicolumn{3}{c}{BIM} &
   &
  \multicolumn{3}{c}{AutoPGD} &
   &
  C\&W \\ \cmidrule{4-6} \cmidrule{8-10} \cmidrule{12-14} \cmidrule{16-18}
 Perturbation size $\epsilon$ &      & -     & 0.003 & 0.01  & 0.03  &  & 0.003 & 0.01  & 0.03  &  & 0.003 & 0.01  & 0.03  &  & 0.003 & 0.01  & 0.03  &  & -     \\ \midrule

\multirow{2}{*}{Chest X-ray} &
  ViT &
  \textbf{96.38} &
  91.59 &
  77.39 &
  55.65 &
   &
  92.17 &
  69.86 &
  32.32 &
   &
  91.01 &
  66.38 &
  28.99 &
   &
  92.64 &
  63.77 &
  21.30 &
   &
  47.83 \\
           & SEViT & 94.64 & 94.20 & 92.03 & 89.28 &  & 94.35 & 92.61 & 88.41 &  & 94.06 & 91.88 & 86.67 &  & 94.20 & 92.17 & 86.51 &  & 93.62 \\ \midrule
Fundoscopy & ViT  & \textbf{97.64} & 35.42 & 17.92 & 16.53 &  & 1.40  & 0.0   & 0.0   &  & 0.30  & 0.0   & 0.0   &  & 0.0   & 0.0   & 0.0   &  & 9.00  \\
           & SEViT & 85.62 & 80.69 & 80.69 & 80.97 &  & 78.06 & 77.40 & 77.64 &  & 77.50 & 77.78 & 77.06 &  & 77.36 & 78.33 & 79.03 &  & 78.33 \\ \bottomrule
\end{tabular}%
}
\end{table}
\begin{table}[t]
\centering
\caption{Classification accuracy (\%) of SEViT based on majority voting from $c$ randomly selected intermediate classifiers along with the final ViT classifier. All reported results are obtained by averaging over 5 trials.}
\label{Random_Both_datasets}
\resizebox{0.75\textwidth}{!}{%
\begin{tabular}{@{}clllccccccccclccccccc@{}}
\toprule
                         &  &                    &                   & \multicolumn{1}{l}{} & \multicolumn{7}{c}{Chest X-ray}        & \multicolumn{1}{l}{} &  & \multicolumn{7}{c}{Fundoscopy}         \\ \cmidrule{5-12} \cmidrule{15-21}
Number  of  MLPs         &  & \multicolumn{2}{l}{}                   &                      & 1     &  & 2     &  & 3     &  & 4     &                      &  & 1     &  & 2     &  & 3     &  & 4     \\
\midrule
Clean Samples            &  & \multicolumn{2}{c}{$ -$}               &                      & \textbf{94.25} &  & \textbf{95.26} &  &\textbf{ 94.72} &  & \textbf{95.30} &                      &  & \textbf{84.30} &  & \textbf{87.87} &  & \textbf{85.50} &  & \textbf{85.40} \\
\multirow{3}{*}{FGSM}    &  & \multicolumn{2}{l}{$\epsilon = 0.003$} &                      & 91.62 &  & 94.29 &  & 93.62 &  & 94.84 &                      &  & 69.89 &  & 78.42 &  & 78.33 &  & 81.31 \\
                         &  & \multicolumn{2}{l}{$\epsilon = 0.01$}  &                      & 83.54 &  & 91.36 &  & 91.40 &  & 92.32 &                      &  & 57.14 &  & 81.11 &  & 80.75 &  & 87.97 \\
                         &  & \multicolumn{2}{l}{$\epsilon = 0.03$}  &                      & 72.84 &  & 88.06 &  & 88.10 &  & 89.28 &                      &  & 55.06 &  & 83.81 &  & 83.67 &  & 81.92 \\
                         \cmidrule{1-4}
\multirow{3}{*}{PGD}     &  & \multicolumn{2}{l}{$\epsilon = 0.003$} &                      & 92.32 &  & 94.26 &  & 93.77 &  & 94.87 &                      &  & 44.06 &  & 71.36 &  & 72.67 &  & 78.94 \\
                         &  & \multicolumn{2}{l}{$\epsilon = 0.01$}  &                      & 83.39 &  & 90.93 &  & 91.07 &  & 92.35 &                      &  & 42.08 &  & 74.75 &  & 73.28 &  & 79.42 \\
                         &  & \multicolumn{2}{l}{$\epsilon = 0.03$}  &                      & 66.20 &  & 85.94 &  & 87.51 &  & 88.26 &                      &  & 42.06 &  & 78.64 &  & 78.01 &  & 79.50 \\
                         \cmidrule{1-4}
\multirow{3}{*}{BIM}     &  & \multicolumn{2}{l}{$\epsilon = 0.003$} &                      & 91.51 &  & 93.97 &  & 93.48 &  & 94.58 &                      &  & 42.83 &  & 70.94 &  & 72.25 &  & 78.78 \\
                         &  & \multicolumn{2}{l}{$\epsilon = 0.01$}  &                      & 81.07 &  & 90.46 &  & 90.87 &  & 91.86 &                      &  & 41.94 &  & 75.72 &  & 75.11 &  & 79.31 \\
                         &  & \multicolumn{2}{l}{$\epsilon = 0.03$}  & \multicolumn{1}{l}{} & 62.35 &  & 84.49 &  & 85.16 &  & 86.75 &                      &  & 41.72 &  & 79.63 &  & 79.01 &  & 79.50 \\
                         \cmidrule{1-4}
\multirow{3}{*}{AutoPGD} &  & \multicolumn{2}{l}{$\epsilon = 0.003$} & \multicolumn{1}{l}{} & 92.55 &  & 94.46 &  & 93.83 &  & 94.89 &                      &  & 42.43 &  & 70.83 &  & 72.11 &  & 78.64 \\
                         &  & \multicolumn{2}{l}{$\epsilon = 0.01$}  & \multicolumn{1}{l}{} & 81.10 &  & 90.49 &  & 90.75 &  & 92.14 &                      &  & 42.06 &  & 76.50 &  & 76.31 &  & 80.11 \\
                         &  & \multicolumn{2}{l}{$\epsilon = 0.03$}  & \multicolumn{1}{l}{} & 60.1  &  & 84.06 &  & 85.28 &  & 87.30 &                      &  & 42.06 &  & 79.89 &  & 76.20 &  & 79.70 \\
                         \cmidrule{1-4}
C\&W                     &  & \multicolumn{2}{c}{$-$}                & \multicolumn{1}{l}{} & 69.33 &  & 92.26 &  & 93.70 &  & 93.68 &                      &  & 48.53 &  & 74.89 &  & 76.69 &  & 80.97 \\ \bottomrule
\end{tabular}%
}
\end{table}
For SEViT models with a random selection of $c$ classifiers instead of choosing all the $m$ intermediate classifiers, the accuracy for different values of $c$ are reported in Table~\ref{Random_Both_datasets}. The results show that SEViT has reasonable robust accuracy even with $c=1$. However, as $c$ increases to $4$, the robust accuracy is close to that of the full SEViT model with $m=5$. These results validate our hypothesis that patch tokens from initial ViT blocks can be utilized along with the class token of the last ViT block to enhance the robustness of the model.
\subsection{Results on Adversarial Sample Detection} 
\begin{figure}[t!]
\centering
\includegraphics[width=0.82\textwidth, keepaspectratio]{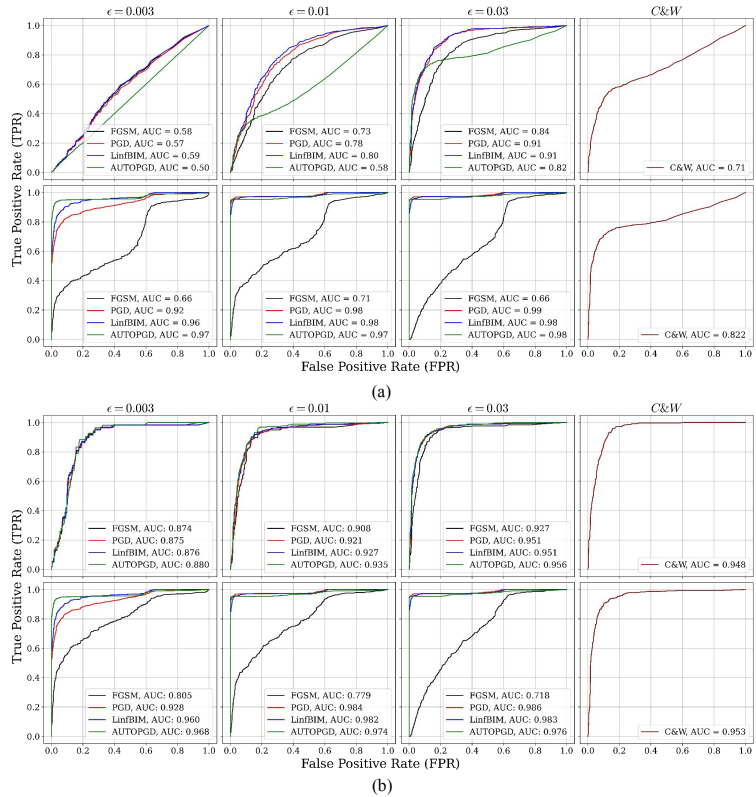}
\caption{ROC curves along with AUC values for (a) all adversarial samples (b) adversarial samples that succeeded in fooling vanilla ViT. First row in each sub graph corresponds to Chest X-ray dataset, the second row corresponds to Fundoscopy dataset.}
\label{ROC_total_old}
\end{figure}

We evaluate the ability of SEViT to detect adversarial samples, regardless of the success of these attacks in fooling the vanilla ViT. Fig.~\ref{ROC_total_old} (a) show the ROC curves along with the Area Under ROC (AUC). SEViT detects attacks with larger perturbation budget more effectively for both datasets. However, the detection performance is low for smaller perturbation budget and for FGSM attacks on Fundoscopy images. This is mainly because attacks with smaller perturbation budget introduce only negligible changes to the input images. This is confirmed by the inability of such samples to fool the vanilla ViT. Considering only adversarial samples that were successful in fooling the ViT, SEViT detects PGD, BIM, AutoPGD and C\&W attacks with $0.93$ average AUC, whereas the average AUC is $0.76$ for detecting FGSM attacks on Fundoscopy images (Fig. \ref{ROC_total_old} (b)). 

\section{Conclusion}
In this paper, we proposed a novel Self-Ensemble Vision Transformer (SEViT) architecture, which leverages the feature representations learned by initial blocks of ViT to train intermediate classifiers for medical imaging classification. The predictions from intermediate and final ViT classifiers are combined to enhance the robustness against adversarial attacks, and the consistency of the predictions to detect the adversarial samples. We prove the effectiveness of SEViT using two different publicly available medical datasets from different modalities. In the future, we aim to (i) extend our work and evaluate SEViT against Transformer-based attacks and under a full white-box setting where the adversary has complete knowledge about SEViT architecture and (ii) evaluate SEViT in the context of natural images.

\bibliographystyle{splncs04}
\bibliography{paper1676}
\end{document}